\documentclass[conference]{IEEEtran}
\IEEEoverridecommandlockouts
\usepackage{cite}
\usepackage{amsmath,amssymb,amsfonts}
\usepackage{algorithmic}
\usepackage{graphicx}
\usepackage{textcomp}
\usepackage{xcolor}
\usepackage{hyperref}
\def\BibTeX{{\rm B\kern-.05em{\sc i\kern-.025em b}\kern-.08em
    T\kern-.1667em\lower.7ex\hbox{E}\kern-.125emX}}
\begin{document}

\title{Recognising Known Configurations of Garments For Dual-Arm Robotic Flattening\\
}

\author{\IEEEauthorblockN{Li Duan}
\IEEEauthorblockA{\textit{School of Computing Science} \\
\textit{University of Glasgow}\\
Glasgow, United Kingdom\\
l.duan.1@research.gla.ac.uk}
\and
\IEEEauthorblockN{Gerardo Aragon-Camarasa}
\IEEEauthorblockA{\textit{School of Computing Science} \\
\textit{University of Glasgow}\\
Glasgow, United Kingdom \\
gerardo.aragoncamarasa@glasgow.ac.uk}
}

\maketitle

\title{Recognising Known Configurations of Garments For Dual-Arm Robotic Flattening\\
}

\author{\IEEEauthorblockN{Li Duan}
\IEEEauthorblockA{\textit{School of Computing Science} \\
\textit{University of Glasgow}\\
Glasgow, United Kingdom\\
l.duan.1@research.gla.ac.uk}
\and
\IEEEauthorblockN{Gerardo Aragon-Camarasa}
\IEEEauthorblockA{\textit{School of Computing Science} \\
\textit{University of Glasgow}\\
Glasgow, United Kingdom \\
gerardo.aragoncamarasa@glasgow.ac.uk}
}

\maketitle

\begin{abstract}
Robotic deformable-object manipulation is a challenge in the robotic industry because deformable objects have complicated and various object states. Predicting those object states and updating manipulation planning is time-consuming and computationally expensive. In this paper, we propose learning \textit{known configurations} of garments to allow a robot to recognise garment states and choose a pre-designed manipulation plan for garment flattening.
\end{abstract}

\begin{IEEEkeywords}
pattern recognition; robot vision; depth images; deformable objects
\end{IEEEkeywords}

\section{Introduction \label{sec:introduction}}
Robots have been widely applied in many aspects of life, for example in robotic assistance for humans, life-threatening rescue, and manufacturing. Among these applications, robotic deformable object manipulation plays a critical part. Autonomous garment laundry and sorting, soft-object manipulation in life-threatening rescue and manufacturing, and robotic threading in the textile industry require an efficient and robust robotic deformable-object manipulation strategy. 

However, robotic manipulation of deformable objects remains an open problem in robotics because deformable objects can take unpredictable object states (crumpled, stretched, or bent configurations). That is, their states frequently vary when they are being manipulated and can not be predicted precisely and timely. Deformable objects are easily slipped from robots and take a long time to be manipulated because of their properties. Therefore, current research efforts focus on reducing object-state or robotic action-state complexity.

Current approaches that reduce deformable object complexity for robotic manipulation can be divided into two categories: those that propose to find simplified models to represent object states or action states first~\cite{yan2020learning,li2015folding,10.1145/182478.182583,zaidi2017model,7139370,8255664}, and those that rely on model-free reinforcement learning solutions where the agent learns to model the object while interacting with it~\cite{pignat2017learning,schulman2016learning,balaguer2011combining}. When deformable object models are simplified, it is possible to reduce the computational costs, making it possible to predict the object states of manipulated deformable objects close to real-time. An efficient prediction of future object states enables a robot to update its manipulation plan on-the-fly, which is critical to manipulating deformable objects in a real-time setting. However, current research is geared towards simple geometries such as ropes, towels or cube-shaped objects. When a reinforcement learning solution is used, data-driven reinforcement learning agent learns manipulation policies using a reward-based mechanism. Robots are rewarded if their manipulation policies produce actions that contribute to a step closer to the object's goal state. However, most of the reinforcement learning approaches rely on simulated environments of which the physics of deformable objects are simplified.

In this paper, we propose taking advantage of gravity to model and learn the \textit{known configurations} of deformable objects (garments in this paper), which can be later used to choose a pre-designed manipulation plan based on the recognised \textit{known configurations} to flatten these objects. Our approach does not need to find simplified models that provide good representations of object states for deformable objects. Similarly, our approach does not require manual action labelling compared to reinforcement-learning approaches. In this paper, we focus on recognising the \textit{known configurations} of garments and develop a pipeline to flatten garments, see Figure \ref{fig:main_fig}.


In summary, our main contributions in this paper are two-fold: we propose learning the \textit{known configurations} of garments, which is the foundation for a full pipeline of an effective robotic garment flattening pipeline. Compared with state-of-art, our approach has a higher accuracy (89\% versus 73\% in \cite{8255664,7139370})
and we captured a \textit{known configurations} database which comprises RGB and depth images of garments

We introduce the related work in section \ref{sec:related_work}, and our methodology in section \ref{sec:methodology}. Our experiment results are in section \ref{sec:experiments_and_results} and a discussion in section \ref{sec:discussion}. Finally, we conclude this paper and suggest future work in section \ref{sec:conclusion_and_future_work}.

\section{Related Work \label{sec:related_work}}
Research on robotic deformable-object manipulation is divided into two categories: model-based approaches and model-free approaches. Model-based approaches focus on engineering methods, simplifying models representing object and action states or applying dynamics modelling rather than state-space modelling. On the other hand, model-free approaches concentrate on devising data-driven methods mainly relating to reinforcement or imitation learning. Robots learn to manipulate deformable objects by applying manipulation policies, usually trained on simulated environments.

The main two approaches of models-based approaches are object-state modelling and action-state modelling. Lin \textit{et al.} \cite{lin2015picking} have proposed to use finite element methods (FEM) to model the dynamics of bottles, where they defined a ‘liftability ’ test. A bottle can only be lifted if its deformation passes the ‘liftability’ test, so the bottle can be stably manipulated without slipping from the robot's gripper. A mass-spring system has been applied in \cite{zaidi2017model} to grasp deformable objects using a multi-fingered robotic hand. The Nvidia PhysX simulator has also been utilised in \cite{yu2017haptic} to teach a robot for assisted human dressing. Miller \textit{et al.}~\cite{miller2012geometric} have proposed a robotic laundry folding approach by introducing a quasi-static cloth model that represents states of garments and an algorithm that outputs a motion planning based on a 2D cloth polygon generated from the quasi-static cloth model and a desired sequence of folds. Yamakawa \textit{et al.}~\cite{yamakawa2012simple} have also proposed learning a simplified model to represent the state space of ropes and derive motion planning from the inverse dynamics of rope states. McConachie \textit{et al.}\cite{mcconachie2020learning} have suggested learning reduced state spaces by using a classifier to bias a planner away from state-action pairs that are not feasible under general robotic working environments.

The problem with model-based approaches is that due to the high dimensionality of the deformable object and action spaces, only objects with simple geometries (such as towels, ropes, and sponges) can be modelled and manipulated. Objects with complex geometries such as garments can not be used in these approaches because modelling and manipulation are time-consuming and computationally costly. An alternative to simplify model representations is to convert unknown object states to known object states. Li \textit{et al.} \cite{8255664} \cite{7139370} have proposed to estimate the pose of hanging garments and apply manipulation plans based on the recognised garments’ poses. Poses only relate to grasping points rather than initial object states of garments, effectively converting garments from unknown object states to known object states. But their approach requires re-grasping and rotating garments to recognise the poses, adding a time overhead for executing a garment manipulation task.

Data-driven approaches have been recently proposed, namely imitation and reinforcement learning. That is, Pignat \textit{et al.} \cite{pignat2017learning} have proposed to encode sensory information and motor commands as a joint distribution in a hidden semi-Markov model, of which parameters are learned from a set of human demonstrations. Each model set represents a sensorimotor pattern whose sequencing can produce complex behaviours. Balaguer \textit{et al.} \cite{balaguer2011combining} have presented a combined approach of imitation and reinforcement learning, where human demonstrations are used to reduce the search space of the reinforcement learning agent and allow the solution to converge quickly. Tsumine \textit{et al.}~\cite{tsurumine2019deep} have proposed a deep reinforcement learning for robotic cloth manipulation with a smooth policy update. Their networks combine the nature of smooth policy updates in value-function-based reinforcement learning with automatic feature extraction from high-dimensional observations in deep neural networks to enhance the sample efficiency and learning stability with fewer samples. Twardon \textit{et al.} propose restricting robotic action-space modelling to the actions that are safe for robotic hands and arms, garments and the head when a robot is trained to put a knit cap on a Styrofoam head. A direct policy-search algorithm finds appropriate trajectories in the restricted action space in order to enable the robotic knit-cap operation. Their approach has demonstrated an effective manipulation planning under a restricted action space. McConachie \textit{et al.} \cite{mcconachie2018estimating} have proposed formulating a deformable object manipulation task as a multi-armed bandit problem, with each arm representing a model of the deformable object. Matas \textit{et al.} \cite{matas2018sim} use a tailored version of the deep deterministic policy gradients (DDPG \cite{lillicrap2015continuous}) for deformable object manipulation, where they have learned a policy from simulation data while testing the policy on real data. The limitation of imitation and reinforcement learning approaches is that they need several training iterations to learn tasks. Matas \textit{et al.} \cite{matas2018sim} trained a robot to fold a towel and drape a piece of cloth through a hanger by reinforcement learning. The robot is trained in a simulated environment for around 250 iterations. Training such a robot in real environments is difficult because some actions need to be validated in a simulated environment. Hazardous actions (actions that may harm robots or human beings) are easy to handle in simulated environments but need to be avoided in real environments.

\section{Methodology \label{sec:methodology}}
As described in section \ref{sec:introduction}, state-of-art approaches conduct manipulations on tables or platforms, where deformable objects have complicated and various configurations (or object states) and result in computationally costly updates to find a manipulation plan. In contrast, we consider taking advantage of gravity to control the variety of object configurations and reduce the complexity of deformable object configurations.

\begin{figure*}
    \centering
    \includegraphics[width=0.9\textwidth]{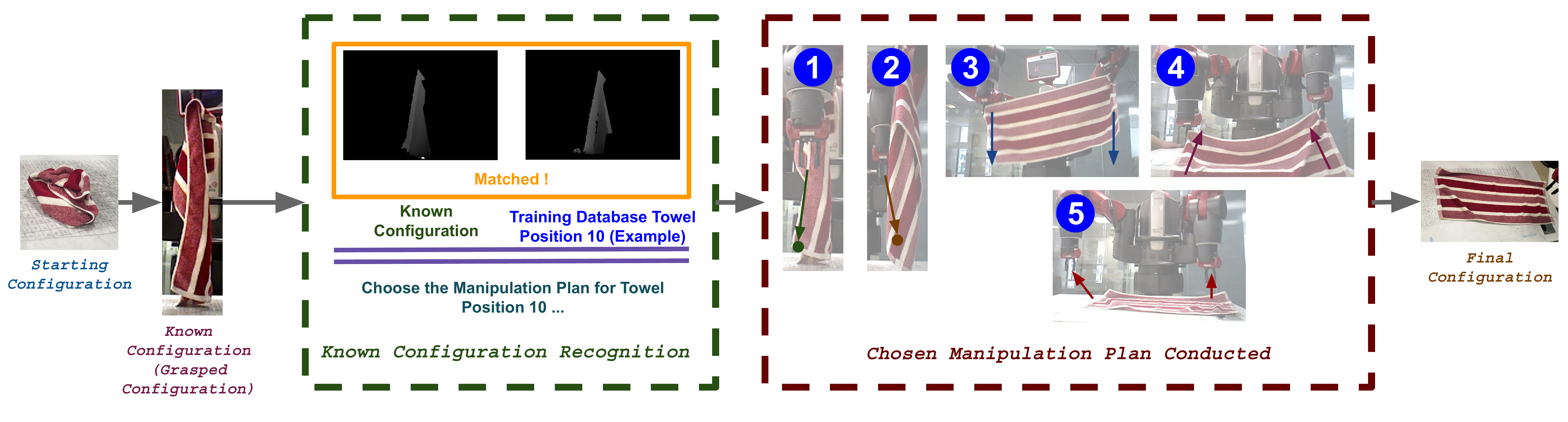}
    \caption{\textit{\textit{Known Configurations} Pipeline}: a towel with a crumpled configuration (starting configuration) is grasped by a robot from a grasping point (the first grasping point as described in Section \ref{sec:methodology}), giving rise to a \textit{known configuration}. The \textit{known configuration} is recognised, and a pre-designed five-step manipulation plan (described in section \ref{sec:methodology}) can be chosen based on the recognised \textit{known configuration}. The robot manipulates the towel with the chosen manipulation plan.}
    \label{fig:main_fig}
\end{figure*}

In this paper, we start by assuming that garments of the same categories (e.g. jeans, towels, tshirts, etc.) lying on a table have different configurations. If the garments are grasped from similar grasping points, they will have similar configurations when the robot picks them up, which we call \textit{known configurations} in this paper. For example two towels have different crumpled configurations on a table (i.e. starting configurations), but they have similar \textit{known configurations} after a robot grasps them. These \textit{known configurations} only depend on the grasping points because of gravity. That is, we convert complex starting configurations to simple \textit{known configurations} from which the robot can follow pre-defined manipulation plans to flatten them. Recognising the \textit{known configurations} of garments is crucial for the robotic garment flattening pipeline shown in Figure \ref{fig:main_fig}.

After the garment's \textit{known configuration} is obtained after grasping them from the table, we find the second grasping points based on how gravity effect on the garment. For example, as shown in Figure \ref{fig:main_fig} step 1 in the \textit{manipulation routine} box, the second grasping point is the lowest point of the towel. Different \textit{known configurations}  have different locations of the second grasping points, so recognising the garment's \textit{known configurations} are critical to localise the second grasping point. After the robot finds the second grasping point, the next step (step two in Figure \ref{fig:main_fig}) is to find the third grasping point, which is the opposing ending corner of the towel. Then, the robot stretches the towel from the two grasping points in step three in Figure \ref{fig:main_fig}. Because of the stretching and gravity, the towel is in flattened state; therefore, the final step consists of placing the towel on the table by sliding it on the edge of the table (shown in steps four and five in Figure \ref{fig:main_fig}).


To recognise garment's \textit{known configurations}, we propose to use an off-the-shelf deep neural network (named KCNet in this paper) based on ResNet-18\cite{he2016deep}. KCNet is mathematically expressed as: $O=F(C(I))$, where $O$ is the output of the KCNet, which is the recognised \textit{known configurations} (represented as classification class), $I$ is the input image that captures \textit{known configurations}, $C$ are convolutional layers (a ResNet-18\cite{he2016deep}), and $F$ are fully connected layers. We use a negative log-likelihood loss (NLLLoss) to train this network: $L(\theta)=\sum_{i=1}^{n}(y_{i}\log{\hat{y}_{\theta,i}}+(1-y_{i})\log(1-\hat{y}_{\theta,i}))$, where $\theta$ is the weight parameters of the network, $y_{i}$ is the ground truth probability that the $i$th data point is positive, and $\hat{y}_{\theta,i}$ is the predicted probability that the $i$th data point is positive. Figure \ref{fig:main_fig} shows how KCNet works in the \textit{known configurations} pipeline.

As stated in section \ref{sec:methodology}, a \textit{known configuration} highly depends on the grasping points, therefore we label \textit{known configurations} in terms to its corresponding grasp points. \textit{Known configurations} should be recognisable by KCNet, so they should be distinguishable from each other. Therefore, a \textit{known configuration} represents all grasping points in a segmented area of a garment, rather than a single grasping point. We proposed to discritise garments into different segments and define a grasping point to represent one segment. 
Our experiments found that dividing a garment into ten segments can provide the most \textit{known configurations} that are recognisable. In each segment, we locate its corresponding grasping point as the centre of the area.


State-of-art focuses on real-time policy updating in robotic deformable object manipulation. A robot will update its manipulation plan after observing a new object state. The updates are computationally intensive and time-consuming. In contrast, we propose designing manipulation plans in advance to avoid updating the manipulation plan to speed up the task execution. 

A manipulation plan comprises sequences of 6D poses and the the robot's gripper state (i.e. open or close). As described in section \ref{sec:methodology}, our manipulation strategy include finding the second and third grasping points, stretching to flatten garments and lifting garments down to the table. Our manipulation plan design is thus based on these strategy and ensures that each step requires the least actions. We defined these sequences in CSV files to transfer manipulation commands to the robot after a \textit{known configuration} is recognised.

\section{Experiments And Results \label{sec:experiments_and_results}}

\subsection{Experimental Methodology \label{sec:experiment_settings}}

Our \textit{known configurations} pipeline consists of three stages: the robot recognises the \textit{known configuration} of a grasped garment, based on the recognised \textit{known configuration}, the robot chooses a pre-defined manipulation plan, and then the robot flattens the garment.

\begin{figure*}[tpbh]
    \centering
    \includegraphics[width=0.8\textwidth]{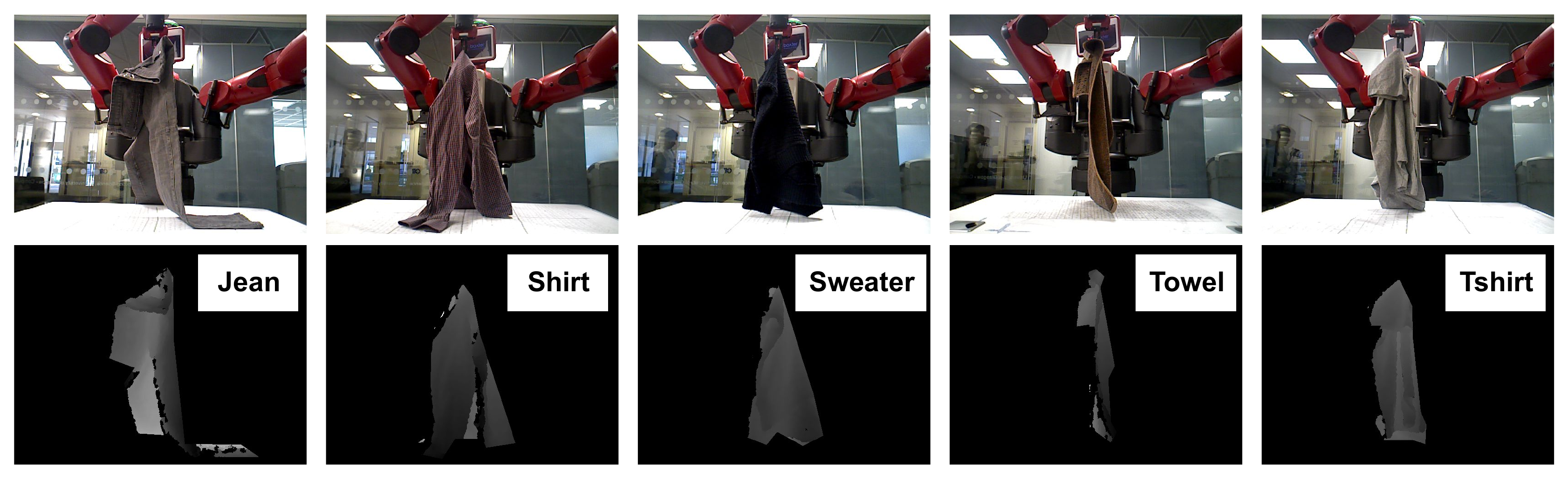}
    \caption{\textit{Database Examples}: There are five garment categories in our database: jeans, shirts, sweaters, towels and tshirts. Our database contains masked depth and RGB images of four garment instances in each category, which have five categories (Top: original images, Bottom: masked depth images).}
    \label{fig:database_examples}
\end{figure*}

To recognise garments' \textit{known configurations}, we captured a novel dataset comprising depth and RGB images of garments from five categories: jeans, shirts, sweaters, towels and tshirts. There are four garment instances in each category, and as described in section \ref{sec:methodology}, each garment instance has ten grasping segments represented by ten grasping positions. We collected 100 images for each position, where each image captures the \textit{known configuration} of a garment grasped from a specific position. We captured a total of 19,269 depth and RGB images, respectively, with a resolution of $256\times 256$ pixels. Figure \ref{fig:database_examples} shows examples of images in this dataset. For each garment category, we pre-defined 10 manipulation plans for each the 10 segments as described in section \ref{sec:methodology}. Therefore, we have 50 manipulation plans.

With this dataset, we trained a KCNet which is a classification network that consists of a pre-trained ResNet 18 structure and a fully connected network. The detailed implementation of KCNet can be found at \url{https://liduanatglasgow.github.io/known_configurations/}. The learning rate is set to $10^{-3}$ and we allow it to decay during the training with a step scheduler (a decay factor of 0.1 and a step size of 8). 

We have implemented a k-fold cross-validation approach (k-fold CVA) to train and test KCNet, rather than the traditional approach of train-validate-test splits. The $k$ value in our experiment is set to four, which means that our database is split into four groups for the k-fold CVA training and testing sessions. There are four garment instances in a group. Three groups are assigned as training groups for each session and one group as a testing group. We ensured that the garment instances in the testing group were ‘unseen’ by KCNet. We iterate the testing group to include all garment instances in a category, and we averaged the classification accuracies as the classification accuracy for the network.

The robot in this experiment is a Baxter dual-arm robot with a table at the front to place the garments. We use a computer with Ubuntu 16.04 and an NVIDIA 1080 Ti GPU to train the KCNet. The robot is controlled by the Robot Operating System, and an Xtion camera captures the images. In section \ref{subsec:results_and_demonstration}, we demonstrate an example of a five-step manipulation plan based on the recognised \textit{known configurations} of a towel for flattening the towel. We will improve pre-designed manipulation plans in our future research.

\subsection{KCNet Validation Results and Manipulation Demonstration \label{subsec:results_and_demonstration}}
We compared the performance of KCNets trained on RGB, depth and RGBD images. Table \ref{tab:result_comparison} shows our results for each category in each k-fold cross-validation described in section \ref{sec:experiment_settings}.  Table \ref{tab:result_comparison} shows classification accuracies that represent the percentages of \textit{known configurations} recognised by the KCNet.

We found that training a KCNet on depth images (89\%) outperforms a KCNet on RGB images (73\%). Depth images capture structural and physical characteristics of garments (as found in \cite{duan2022physnet}), and therefore, enables a KCNet to recognise the \textit{known configurations} of unseen garment instances. The RGB images capture texture characteristics that depend on the lighting condition and shadows which hinders the ability of KCNet to predict \textit{known configurations}. Meanwhile, We can observe an increase from 73\% to 78.5\% for the KCNet trained on RGBD images but still a significant gap behind the KCNet trained only on depth images (89.0\%). Depth information facilitates a KCNet to recognise the \textit{known configurations} of unseen garments, while RGB information may affect the effectiveness of depth information.

\begin{figure*}[thbp]
    \centering
    \includegraphics[width=0.9\textwidth]{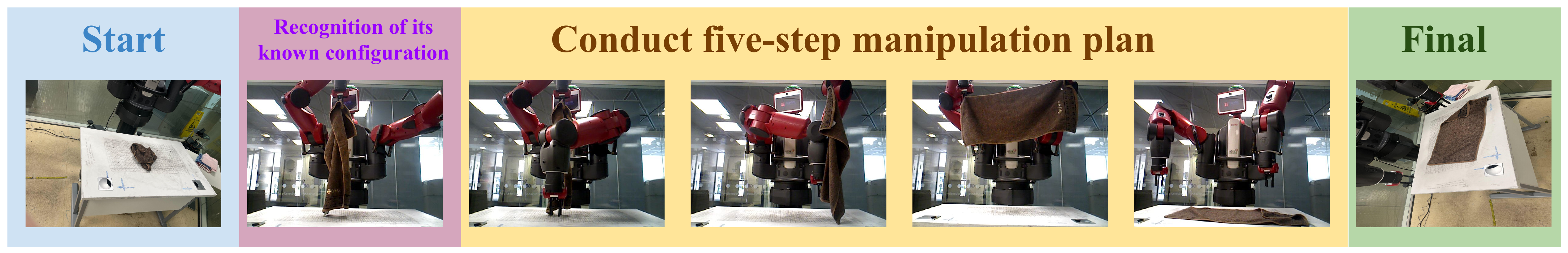}
    \caption{\textit{An Example of Manipulations}: an example are demonstrated: firstly, the robot grasps the towel from the table to recognise its \textit{known configurations}. Then a manipulation plan is chosen. Finally, the robot flattens the towel with the chosen manipulation plan.}
    \label{fig:examples_manipulation}
\end{figure*}

Based on this ablation study, we chose the KCNet trained on the depth images to conduct the manipulation experiments. Figure \ref{fig:examples_manipulation} shows an example of manipulations in our experiments. As described in section \ref{sec:methodology}, the robot firstly recognises the \textit{known configurations} of the garments, then chooses a manipulation plan based on the recognised \textit{known configurations}, and finally flattens the garments with the chosen manipulation plan. We can observe that the towel shown in Figure \ref{fig:examples_manipulation} are correctly flattened from its crumpled configuration. A robotic manipulation example can be found on our project website \url{https://liduanatglasgow.github.io/known_configurations/}.

\begin{table*}[t]
    \centering
    \caption{k-fold cross validation experiment results: comparison between depth, RGB and RGBD images (unit: \%)}
    \label{tab:result_comparison}
    \begin{tabular}{|c|c|c|c|c||c|c|c|c|c||c|c|c|c|c|}
    \hline
     \textbf{Depth}  & 1 & 2 & 3 & 4 & \textbf{RGB} & 1 & 2 & 3 & 4 &\textbf{RGBD}  & 1 & 2 & 3 & 4\\ \hline
     towel &94.4&96.4&93.1& 86.2& towel & 67.0& 55.9& 74.1& 64.5 &towel & 71.8 & 79.7 & 85.2 & 76.7\\ \hline
     tshirt &86.8& 87.2&96.3 & 94.7 & tshirt & 70.8& 71.2& 72.1& 76.9 &tshirt & 74.6 & 68.3 & 84.4 & 83.9\\ \hline
     shirt &78.2& 80.3&75.9 & 92.5 & shirt & 87.9& 58.4& 75.4& 91.4 &shirt & 87.9 & 58.4 & 68.8 & 91.9\\ \hline
     sweater &78.4& 85.4 &87.0 &86.2& sweater & 54.6& 42.0& 68.1& 85.2 &sweater & 52.7 & 51.6 & 74.7 & 77.0\\ \hline
     jean &99.3 & 95.8& 95.3&99.1& jean & 76.4&87.8 & 87.0& 98.8 &jean & 87.5 & 92.0 & 97.2 & 99.5\\ \hline
     \textit{average} & \textit{87.0} & \textit{89.0} & \textit{89.0}& \textit{92.0} & \textit{average} &\textit{73.0}& \textit{62.0} & \textit{75.0} & \textit{83.0} &\textit{average} & \textit{76.0} & \textit{70.0} & \textit{82.0}& \textit{86.0} \\ \hline
     \multicolumn{5}{|c|}{AVERAGE: \textit{\textbf{89.0}}}&  \multicolumn{5}{|c|}{AVERAGE: \textbf{73.0}} & \multicolumn{5}{|c|}{AVERAGE: \textbf{78.5}}\\
     \hline
    \end{tabular}
\end{table*}

\section{Discussion \label{sec:discussion}}
{red}{In our experiments, we proposed a robotic garment manipulation by recognising garments’ \textit{known configurations} and choosing a pre-designed manipulation plans to flatten garments. This paper introduces the first part of this pipeline: how to recognise the ‘known configurations’ of garments.  The \textit{known configurations} can be recognised because we took advantage of gravity and reduced deformable objects' complexity. Various and complicated initial configurations are converted into simple \textit{known configurations} by taking advantage of gravity after the robot has grasped the garment.} 



\section{Conclusion And Future Work \label{sec:conclusion_and_future_work}}
We proposed an effective robotic garment flattening approach by recognising the \textit{known configurations} of garments and choosing a pre-designed manipulation plan to flatten them. Our approach also features training KCNets on depth images of real garments, resulting in higher recognition accuracy compared with previous work \cite{8255664}.

In our experiments, however, a robot can only recognise the \textit{known configurations} of the garments with shapes from the five categories, while it is unable to recognise those of the garments with an unknown category. Our future work aims to devise a continual learning framework so that a robot learns unknown shapes when it deals with those shapes. Additionally, we believe that a robot can benefit from prior knowledge of shapes \cite{duan2022garnet} or physics properties \cite{duan2022physnet} of garments to recognise \textit{known configurations}, so we propose learning that prior knowledge during the robot’s grasping garments and improving recognition accuracy on \textit{known configurations}.

\section{Acknowledgement \label{sec:acknowledgement}}
We thank the members of the Computer Vision and Automation System group from the University of Glasgow for discussing the experimental results and providing experiment suggestions.

\bibliographystyle{IEEEtran}
\bibliography{references.bib}
\end{document}